\pdfoutput=1

\documentclass[11pt]{article}

\usepackage[preprint]{acl}

\usepackage{times}
\usepackage{latexsym}
\usepackage{stfloats}
\usepackage{hyperref}

\usepackage[T1]{fontenc}

\usepackage[utf8]{inputenc}

\usepackage{microtype}

\usepackage{inconsolata}

\usepackage{graphicx}
\usepackage{caption}
\usepackage{subcaption}
\usepackage[normalem]{ulem}
\usepackage{amsmath, bm}
\usepackage{amssymb}

\title{Enhancing Spoken Discourse Modeling in Language Models Using Gestural Cues}

\author{
 \textbf{Varsha Suresh\textsuperscript{1}}, \textbf{M. Hamza Mughal\textsuperscript{2}}, \textbf{Christian Theobalt\textsuperscript{1,2}}, \textbf{Vera Demberg\textsuperscript{1,2}}
\\
\\
 \textsuperscript{1}Saarland University,
 \textsuperscript{2}Max Planck Institute for Informatics, Saarland Informatics Campus
\\
\small{
\textbf{Correspondence:}
 \texttt{\{vsuresh,vera\}@lst.uni-saarland.de}},
 \small{\texttt{\{mmughal,theobalt\}@mpi-inf.mpg.de}}
}

\begin{document}
\maketitle
\begin{abstract}
Research in linguistics shows that non-verbal cues, such as gestures, play a crucial role in spoken discourse. For example, speakers perform hand gestures to indicate topic shifts, helping listeners identify transitions in discourse. In this work, we investigate whether the joint modeling of gestures using human motion sequences and language can improve spoken discourse modeling in language models. To integrate gestures into language models, we first encode 3D human motion sequences into discrete gesture tokens using a VQ-VAE. These gesture token embeddings are then aligned with text embeddings through feature alignment, mapping them into the text embedding space. To evaluate the gesture-aligned language model on spoken discourse, we construct text infilling tasks targeting three key discourse cues grounded in linguistic research: discourse connectives, stance markers, and quantifiers. Results show that incorporating gestures enhances marker prediction accuracy across the three tasks, highlighting the complementary information that gestures can offer in modeling spoken discourse. We view this work as an initial step toward leveraging non-verbal cues to advance spoken language modeling in language models.
\end{abstract}

\section{Introduction} 
Non-verbal cues, such as gestures, eye contact, body posture, and facial expressions, play a crucial role in communication, functioning as a secondary channel alongside language \cite{krauss1996nonverbal, fichten1992verbal, mcneill1992hand}.
Linguistic research highlights the importance of gestures in spoken discourse \cite{goldin1993transitions, khosrobeigi2022gesture, kita2017gestures, holler2009iconic, cassell2001non}, with studies showing that they help structure conversations, often marking topic shifts \cite{kendon1995gestures, quek2002multimodal}, and convey the speaker's attitudes and certainty about their propositions \cite{roseano2016communicating, andries2023multimodal, debras2015stance}.

Most language models dealing with spoken language data, however, rely solely on textual content and overlook the rich information conveyed through gestural cues. We believe that incorporating this additional non-verbal modality will enhance the language model's capabilities of modeling spoken discourse. Several studies have explored combining modalities such as image and speech with language models \cite{tangsalmonn,tsimpoukelli2021multimodal,team2024chameleon,wangsimvlm}. However, incorporating gesture modality into language modeling remains underexplored. A recent work has taken steps in this direction by defining gesture tokens by encoding the spatial position of a speaker's wrists using coarse grid tokenization on each 2D frame \cite{xu2023spontaneous}. However, gestures often contain more fine-grained information involving coordinated movements of the wrist and shoulder joints. 

In this work, we aim to integrate gesture motion sequences into language models and assess their impact on spoken discourse modeling in language models. Given 3D human motion sequences and corresponding spoken text, we first identify atomic units (tokens) of gestures. Inspired by recent works on co-speech gesture synthesis, we use a VQ-VAE-based architecture to learn these gesture tokens \cite{liu2024emage}. These gesture representations capture fine-grained motion details as they are trained to reconstruct 3D human motion. Once these tokens are obtained, we perform feature alignment to map the feature spaces of the gesture token embeddings to the input space of the language model. This alignment is achieved by training a projection via a joint masked gesture prediction and masked language modeling objective. Finally, we fine-tune the gesture-aligned language model for discourse-based tasks using low-rank adaptation \cite{hu2022lora}.

To evaluate whether the proposed method of incorporating gestures into language models enhances spoken discourse modeling, we propose three text infilling tasks based on key markers in spoken discourse: (i) Discourse Connectives \cite{kendon1995gestures,quek2002multimodal}, (ii) Quantifiers \cite{lorson2024gesture}, and (iii) Stance Markers \cite{roseano2016communicating,andries2023multimodal,debras2015stance}, which linguistic research has shown to frequently co-occur with gestural cues. The text infilling task \cite{wu2019mask}, similar to the Cloze task in linguistics \cite{taylor1953cloze}, involves predicting missing words based on it's surrounding context. In our approach, we mask markers from the three categories during fine-tuning and train the language model to predict them. To achieve this, we follow the connective generation pipeline from \citet{liu2023annotation}. Our experiments demonstrate that incorporating gestures alongside text improves the marker prediction performance across all three tasks. To further validate these findings, we conduct ablation studies and error analysis, offering deeper insights into the model’s performance.

The main contributions of this work are as follows:

\begin{enumerate}

\item We propose a framework that integrates gestural cues into language models and evaluate their potential to improve the language modeling of spoken discourse. 


\item We design three linguistically grounded text infilling tasks to evaluate the performance of spoken discourse modeling.

\item We provide insights on when joint modeling of language and gestures can be helpful based on an in-depth error analysis.

\end{enumerate}

\section{Related Work}

\subsection{Evidence of Co-Speech Gestures from Linguistic Research}

Co-speech gestures are hand and arm movements that temporally co-occur with speech \cite{kendon1972some,mcneill1992hand,shattuck2019dimensionalizing,ebert2024semantics}. 
These gestures can often be broadly classified into two types: rhythmic beat gestures, which are driven by prosody, and semantic gestures, which are context-driven and carry the underlying meaning of the linguistic content 
\cite{mcneill1992hand}. 
%
%
Semantic gestures, in particular, have been shown to frequently co-occur with certain types of spoken discourse markers like discourse connectives, quantifiers and stance markers.

Discourse connectives, such as \emph{however} or \emph{for example}, are used by speakers to signal relationships between different parts of a message. They have been shown to significantly influence gestural patterns associated with lexical markers and across the broader discourse context \cite{mcneill2014discourse}. Gesture features like hand shape or its orientation, help differentiate topics in discourse~\cite{laparle2021discourse,laparle2024more,calbris2011elements, hinnell2019verbal}. 
%
%
For example, raised index fingers are commonly used to express exceptions or concessions (usually signified in language by the words \textit{however} or \textit{but})~\citep{bressem2014repertoire,kendon2004gesture,inbar2022raised}.

%

Stance markers are generally used to express a speaker’s feelings, attitudes, perspectives, or positions \cite{barbara2024corpus, alghazo2021grammatical, schneider2014pragmatics}. Stance can be categorized based on its function in communication. One such category is epistemic stance, which reflects the speaker’s degree of confidence or commitment toward a proposition. Within this category, hedges indicate a reduced level of certainty or commitment, often using words such as \textit{probably} or \textit{may} \cite{wei2021role, deng2023we, liu2021paradigmatic}, whereas boosters strengthen commitment and certainty through expressions like \textit{must}, \textit{really}, and \textit{certainly} \cite{deng2023we, liu2021paradigmatic}. %
Multiple studies have found that speakers frequently use gestures to convey epistemic stance \cite{andries2023multimodal}, such as palm-up gestures to indicate high certainty \cite{marrese2021grammar} and shoulder shrugs for low certainty \cite{roseano2016communicating, borras2011perceiving, debras2015stance}. 

Quantifiers are generally used in spoken discourse to indicate an amount or degree (e.g., \textit{many}, \textit{some}, \textit{all}) and can also express exact numerals \cite{feiman2016logic}. 
\citet{lorson2024gesture} observed that gestural cues help interpret vague quantities like \textit{several}. Their findings suggest that people associate larger horizontal gestures with higher and smaller gestures with lower quantities, indicating a spatio-numerical relationship. 

The findings discussed above highlight the co-occurrence of gestures with respect to three different types of spoken discourse markers. While it is probable that other spoken discourse elements also correlate with gestural cues, our research specifically concentrates on these three types of markers as the primary focus for constructing our text infilling tasks and evaluating the potential of joint modeling of language and gestures.


%
%

%
%
%
%

\subsection{Gesture-Enhanced Language Modeling}

Various works have explored the integration of multiple modalities, such as video, images, and speech, into language models \cite{tsimpoukelli2021multimodal,li2023blip,tangsalmonn,zhang2023speechgpt,hassid2024textually,merullolinearly,liu2024visual}. However, incorporating human motion sequences, which captures gestures, into language modeling is still underexplored.  
As one of the first works in this direction, \citet{abzaliev2022towards} proposed to align gestures with language through CLIP-style embeddings and utilize binary classification to predict the presence of specific word categories like discourse markers.

More recently, there have been efforts to tokenize gestures in a manner similar to language tokens in order to utilize them with transformer-based language models. \citet{xu2023spontaneous} proposed a method to tokenize gestures using coarse grid-based locations of the speaker's hands, employing these tokens for language-modeling tasks.
%
%
%
Similarly, \citet{xu2024llm} explores learning gesture tokens through VQ-VAE~\cite{van2017neural} and combining it with LLMs to perform a `gesture-translation task' for co-speech gesture synthesis. These approaches overlook the rich motion information contained in co-speech gestures. The reliance on coarse grid locations in the former limits its ability to capture language-dependent motion cues like quick hand flicks, while the latter treats gesture tokens as language tokens, and omits motion details contained in the codebook embedding.

To incorporate gesture information into language models, we utilize a gesture encoding process developed for gesture synthesis.
Synthesis methods either represent gestures through continuous latent representations~\cite{mughal2024convofusion, mughal2024raggesture} or tokenized embeddings~\cite{liu2024emage, ao2023gesturediffuclip,xu2024llm}.
The tokenized gesture representations are learned through VQ-VAE based training strategy to capture motion information in the resulting embeddings.
These gesture tokens are suitable to integrate with transformer-based language models and our approach demonstrates the effectiveness of utilizing this information-rich representation of gestures with the language model to improve spoken discourse modeling.

\section{Approach}
\begin{figure*}[h]
\centering
\includegraphics[width=\textwidth]{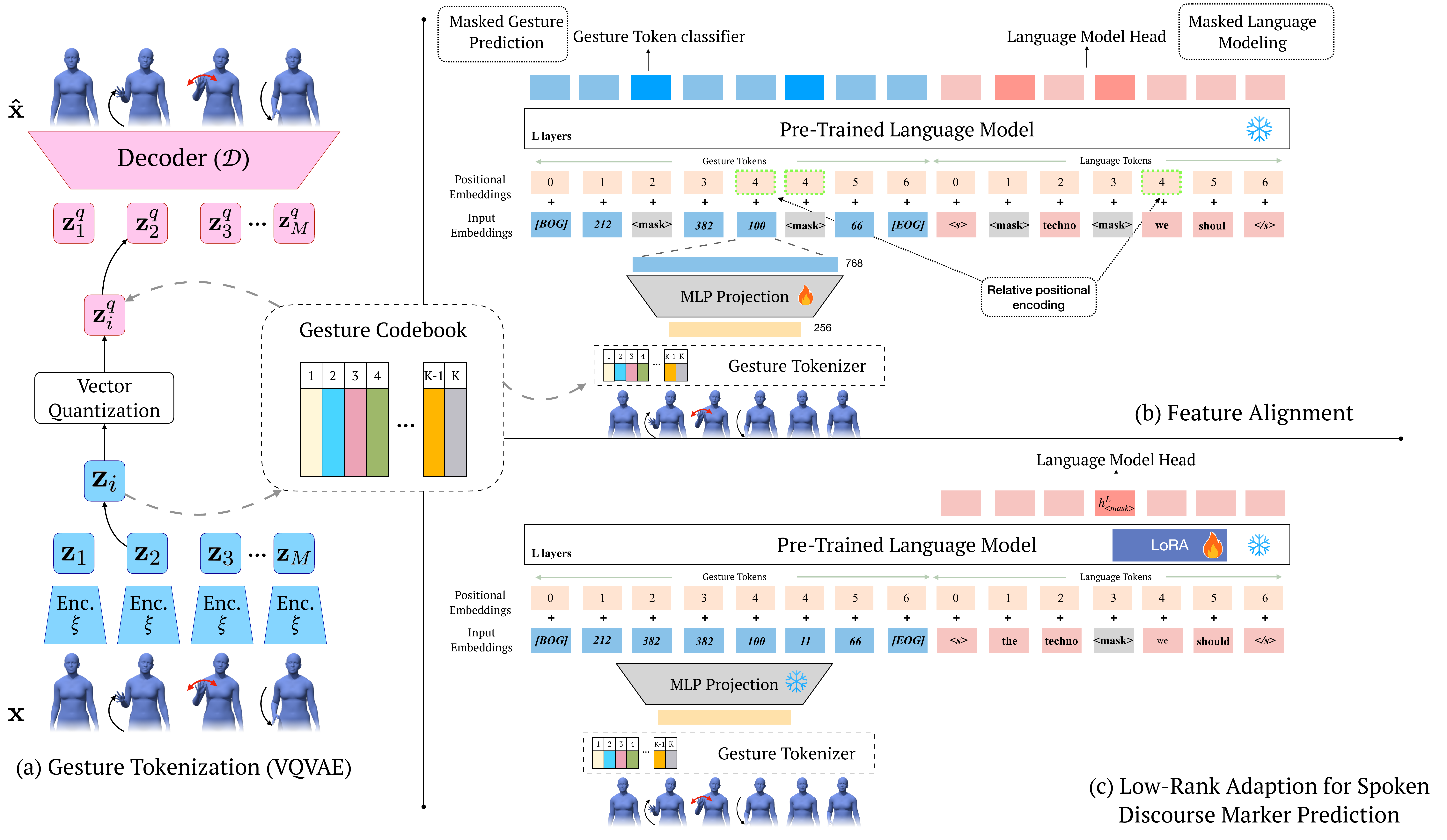}
\vspace{-2em}
\caption{Overall framework of our approach for integrating gestures into language models.}
\label{fig:model}
\end{figure*}

Figure \ref{fig:model} 
illustrates our approach, where the first step involves building the Gesture Tokenizer. This is followed by the feature alignment stage, where the codebook embeddings are aligned with the input embedding space of the language model. The aligned embeddings are subsequently used for fine-tuning on spoken discourse marker prediction tasks.
\subsection{Gesture Tokenization}
Speech-aligned body motion cues provide an information dense signal for downstream language modeling. 
Therefore, we propose to encode this signal into our tokenized gesture representation.
%
%
The gesture sequence $\mathbf{x}$ consists of $N$ frames, with each frame represented by the rotations of $J$ upper body joints.
We convert $\mathbf{x}$ to a tokenized representation $\mathbf{z}^q$ through the VQ-VAE framework~\cite{van2017neural}.
%
%
Given $\mathbf{x}$, we divide the gesture sequence into $M$ chunks and encode each using time-aware transformer encoders ~\cite{mughal2024convofusion}, resulting in a sequence of latent representations  $\mathbf{z} = \{ \mathbf{z}_1,\mathbf{z}_2, ..., \mathbf{z}_M \}$.
Then, in the quantization step, we quantize each latent $\mathbf{z}_i \in \mathbb{R}^d$ to $\mathbf{z}_i^q$ using a codebook of size $K$, which results in the tokenized latent sequence $\mathbf{z}^q = \{ \mathbf{z}_1^q, \mathbf{z}_2^q ..., \mathbf{z}_M^q \}$. 
%
%
Finally, we utilize a single transformer decoder to reconstruct the gesture sequence $\mathbf{\hat{x}}$ from the tokenized latent sequence $\mathbf{z}^q$.
This pipeline is trained on motion reconstruction task where we use standard VQ-VAE losses during training (see Appendix \ref{appendix:gesture_tokenization} for details). 
The resulting Gesture Tokenizer provides an encoding pipeline for the motion cues which can enhance spoken discourse modeling in language models.
%


\subsection{Feature Alignment}
Existing multimodal approaches use various modality fusion architectures, such as cross-attention layers for modality interaction \cite{li2023blip,alayrac2022flamingo}, or unified transformers that tokenize non-textual data (e.g., visual, auditory) to integrate with transformer-based language models \cite{liu2024visual,tsimpoukelli2021multimodal,tangsalmonn,wang2024facegpt}. Our work builds on the latter, with the feature alignment stage aiming to map gesture embeddings into the input embedding space of language models.

For feature alignment, we first obtain paired text and gesture data. For each spoken sentence, we extract the corresponding gesture token embeddings by processing the temporally co-occurring 3D human motion sequence via the gesture tokenizer. We then project the gesture token embeddings using an MLP projector to map them to the input space of the language model. The projected gesture embeddings are concatenated with the input text embeddings and fed into the pre-trained language model, as shown in Figure \ref{fig:model}b. Importantly, only the MLP projector’s parameters are updated during training, while the language model and all other components remain frozen. 

For training, we randomly mask 30\% of both gesture and text tokens and use a joint objective that combines Masked Gesture Prediction ($\mathcal{L}_{\text{MGP}}$) and Masked Language Modeling ($\mathcal{L}_{\text{MLM}}$). For MGP, we train a $K$-class gesture token classifier via Cross-Entropy loss to predict the correct gesture codebook index for masked gesture tokens, similar to MLM.  The final loss for feature alignment ($\mathcal{L}_{\text{FA}}$) is given by, $\mathcal{L}_{\text{FA}} = \mathcal{L}_{\text{MGP}} + \mathcal{L}_{\text{MLM}}$ where

\begin{align*}
\mathcal{L}_{\text{MGP}} &= - \frac{1}{|M_{g}|} \sum_{j \in M_{g}} \log \frac{\exp(x_{y_j})}{\sum_{k \in K} \exp(x_k)} \\
\mathcal{L}_{\text{MLM}} &= - \frac{1}{|M_{t}|} \sum_{i \in M_{t}} \log \frac{\exp(x_{y_i})}{\sum_{v \in V} \exp(x_v)}
\end{align*}




Here $M_{g}$ and $M_{t}$ denotes the sets of masked gesture and text tokens, respectively. The vocabulary size of the language model is represented by $V$ and $K$ denotes the gesture codebook size. The true text and gesture tokens are given by $y_{i}$  and $y_{j}$ respectively and $x$ is the logits.

To ensure temporal alignment between gestures and spoken words, we assign positional embeddings to gesture tokens based on their corresponding text tokens, so that each gesture token’s position matches that of the text token it co-occurs with (as illustrated with the case of token \textit{we} in Figure~\ref{fig:model}). More details regarding computing the gesture positional embedding are provided in the Appendix~\ref{appendix:feature}.

\subsection{Fine-tuning for Spoken Discourse Marker Prediction}

We fine-tune the gesture-aligned pre-trained language model on text infilling tasks targeting three spoken discourse markers: discourse connectives, quantifiers, and stance markers. Details on the label construction for each task are provided in the Evaluation Section~\ref{sec:tasks}. For each marker prediction task we obtain a discrete set of $n$ markers:
$L_{task} = \{l_1,l_2,...,l_{n}\}$ 
where $L_{task}$ represents the subset of the vocabulary relevant to a specific category of markers. We follow the discourse connective generation pipeline outlined in \cite{liu2023annotation}. For this task, the input consists of a sequence of text tokens in the following format:
$ \mathrm{T} =\langle s \rangle \, \mathbf{t_1} \, 
 \langle \text{mask} \rangle  \, \mathbf{t_2} \, \langle /s \rangle$
, where the target marker in the sentence is replaced by $\langle \text{mask} \rangle$ and $\mathbf{t_1}$ and $\mathbf{t_2}$ denote the tokens before and after it. The embedding $h^L_{\langle \text{mask} \rangle}$ obtained after $L$ layers of the transformer model is passed through the Language Modeling Head, similar to MLM. We use the probabilities obtained across the subset vocabulary $L_{task}$ pertaining to the task to perform marker prediction. For fine-tuning we use cross-entropy loss and apply low-rank adaptation \cite{hu2022lora}. As shown in Figure \ref{fig:model}c, we freeze all model components except the adapter layers in the language model during training.


\section{Evaluation}

\subsection{Dataset}

We train and evaluate our method on the BEAT2 dataset~\cite{liu2024emage}, which contains 60 hours of monologue gesture recordings spanning 25 speakers. We use the train/val/test split from the BEAT2 dataset for training the gesture tokenization stage and language modeling stages. 
We use Whisper~\cite{radford2023robust} to obtain the text transcriptions from speech utterances in BEAT2.

\subsection{Tasks}
\label{sec:tasks}
To understand whether incorporating gestures helps in spoken discourse modeling in language models, we construct three text infilling tasks focusing on three lexical markers: discourse connectives, quantifiers, and stance markers. 
To identify the subset of discourse connectives, we process the BEAT2 dataset transcripts using discopy~\cite{knaebel-2021-discopy}, which provides ground truth labels for explicit discourse connectives \footnote{Reported F1 score of 95.6 for explicit connective identification in Section 23 of PDTB 2.0}. In contrast, quantifiers and stance markers are less ambiguous, so we rely on labels sourced from previous studies for both categories (quantifiers: \citealp{madusanka2023not,feiman2016logic,lorson2024gesture,barwise1981generalized}; and stance markers: \citealp{barbara2024corpus,alghazo2021grammatical,schneider2014pragmatics}). Additionally, only markers occurring more than 30 times in the BEAT2 dataset are considered for this study, following the approach of \cite{liu2023annotation}. The labels used in this study are as follows:
\begin{enumerate}
    \item Discourse Connectives: \textit{after, also, although, and, as, because, but, for example, however, if, if then, or, since, so, then, though, when, while}.
    \item Quantifier:  \textit{all, each, enough, entire, few, little, less, many, more, most, much, no, one, some, three, two, whole}.
    \item Stance markers: \textit{actually, almost, amazing, especially, extremely, happy, important, may, maybe, might, must, probably, really, very}. 
\end{enumerate}
The training distribution of these labels are shown below in Figure \ref{fig:data}.

\begin{figure*}[h!]
\centering
\includegraphics[width=\textwidth, clip=TRUE, trim=0cm .35cm 0cm 0cm]{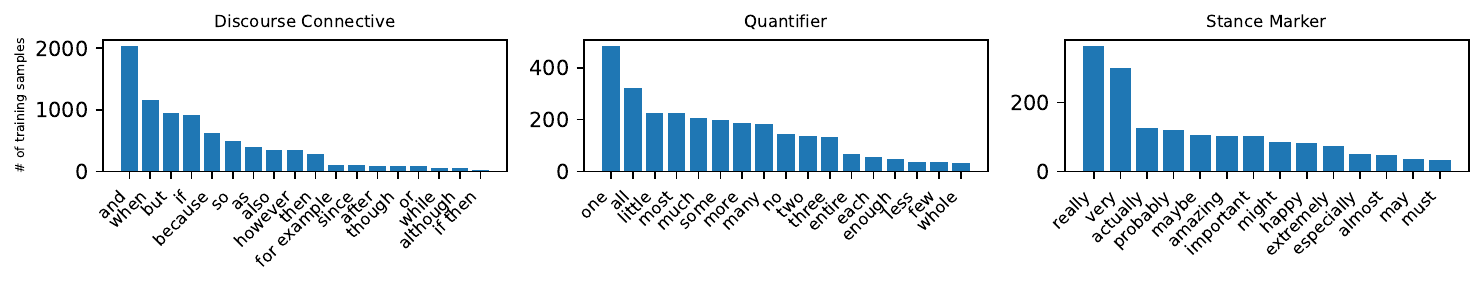}
\caption{Frequency distribution of markers across three tasks: Discourse Connectives, Quantifiers, and Stance Markers in the BEAT2 training data.}
\label{fig:data}
\end{figure*}

\subsection{Implementation Details}

Our approach consists of three main steps: gesture tokenization, feature alignment, and low-rank adaptation. 
To represent the body motion during tokenization, we utilize 6D rotation representation~\cite{zhou2019continuity} for each joint in the upper body, which contains $J=13$ joints.
For the encoding-decoding pipeline, we set the number of motion chunks to $M=8$ per gesture sequence, with each sequence spanning $N=32$ at 15 fps.
The codebook size $K=512$ with embedding length $d=256$ for the gesture tokenizer.
We use the pre-trained RoBERTa-base model \cite{liu2019roberta} as our language model, given its state-of-the-art performance on discourse-based classification tasks \cite{costa2024exploring,zhao2023infusing,zhou2022prompt}. To construct the paired text-gesture data, we introduce special tokens similar to those used in language models, such as $\langle s \rangle$ and $\langle /s \rangle$, to mark gesture sequence boundaries. Specifically, we reserve unused codebook indices for [BOG] (beginning of gesture) and [EOG] (end of gesture), as illustrated in Figure \ref{fig:model}. For feature alignment, we train an MLP projector with two dense layers and GeLU activation. During fine-tuning, we employ LoRA adapters ($r=128, \alpha=256$) for each task while keeping all other model components frozen, as shown in Figure \ref{fig:model}c. All reported results in Tables \ref{tab:main}, \ref{tab:sanity}, and \ref{tab:ablation} are averaged over five random seed runs, with standard deviations included. Further details on hyperparameter choices and training resources are provided in the Appendix \ref{appendix:implementation}.

    

\section{Results}

\begin{table*}[h!]
\resizebox{\textwidth}{!}{%
\begin{tabular}{lcccccc}
\hline
 &
  \multicolumn{2}{c}{\textbf{Discourse}} &
  \multicolumn{2}{c}{\textbf{Quantifer}} &
  \multicolumn{2}{c}{\textbf{Stance}} \\ \hline
 &
  Acc (\%) &
  F1 &
  Acc (\%) &
  F1 &
  Acc (\%) &
  F1 \\ \hline
Text-only baseline &
  60.4 $\pm {\scriptstyle{2.0}}$ &
  47.5 $\pm {\scriptstyle{2.0}}$ &
  69.4 $\pm {\scriptstyle{3.7}}$ &
  65.2 $\pm {\scriptstyle{2.9}}$ &
  50.6 $\pm {\scriptstyle{4.8}}$ &
  46.5 $\pm {\scriptstyle{6.9}}$ \\
\hline
Mixed Modal \citep{xu2023spontaneous} &
  34.8 $\pm {\scriptstyle{1.3}}$ &
  17.4 $\pm {\scriptstyle{0.5}}$ &
  31.7 $\pm {\scriptstyle{0.3}}$ &
  28.2 $\pm {\scriptstyle{0.6}}$ &
  33.4 $\pm {\scriptstyle{2.4}}$ &
  24.0 $\pm {\scriptstyle{3.2}}$ \\ 
GestureLM \small with grid-based tokens \citep{xu2023spontaneous}* &
  55.3 $\pm {\scriptstyle{13.6}}$ &
  41.3 $\pm {\scriptstyle{21.7}}$ &
  70.5 $\pm {\scriptstyle{3.0}}$ &
  65.4 $\pm {\scriptstyle{4.0}}$ &
  47.9 $\pm {\scriptstyle{2.5}}$ &
  44.5 $\pm {\scriptstyle{3.6}}$ \\
GestureLM \small with codebook indices \citep{xu2024llm}* &
  54.4 $\pm {\scriptstyle{13.0}}$ &
  39.2 $\pm {\scriptstyle{20.3}}$ &
  68.4 $\pm {\scriptstyle{2.4}}$ &
  63.9 $\pm {\scriptstyle{2.8}}$ &
  46.5 $\pm {\scriptstyle{6.5}}$ &
  41.7 $\pm {\scriptstyle{5.1}}$ \\
GestureLM (Ours) &
  \textbf{61.2 $\pm {\scriptstyle{1.5}}$} &
  \textbf{51.1 $\pm {\scriptstyle{1.7}}$} &
  \textbf{74.8 $\pm {\scriptstyle{2.6}}$} &
  \textbf{70.4 $\pm {\scriptstyle{3.1}}$} &
\textbf{52.8} $\pm {\scriptstyle{1.7}}$ & \textbf{52.2} $\pm {\scriptstyle{4.3}}$ \\  \hline
\end{tabular}
}
\caption{Comparison with Existing Approaches. The values in the table represent the averaged over five random seed runs, with standard deviations. *ablations to compare different tokenization schemes with our modeling pipeline.}
\label{tab:main}
\end{table*}

\begin{table*}[h]
\centering
\resizebox{0.9\textwidth}{!}{%
\begin{tabular}{lcccccc}
\hline
 & \multicolumn{2}{l}{\textbf{Discourse}} & \multicolumn{2}{l}{\textbf{Quantifer}} & \multicolumn{2}{l}{\textbf{Stance}} \\ \hline
 & Acc (\%)               & F1            & Acc (\%)              & F1          & Acc (\%)              & F1                \\ \hline
Text-only baseline                           & 60.4 $\pm {\scriptstyle{2.0}}$          & 47.5 $\pm {\scriptstyle{2.0}}$          & 69.4 $\pm {\scriptstyle{3.7}}$                           & 65.2 $\pm {\scriptstyle{2.9}}$                           & 50.6 $\pm {\scriptstyle{4.8}}$                           & 46.5 $\pm {\scriptstyle{6.9}}$                           \\ \hline
GestureLM \small with random        & 58.6 $\pm {\scriptstyle{4.4}}$          & 48.1 $\pm {\scriptstyle{3.6}}$          & 73.2 $\pm {\scriptstyle{4.5}}$                           & 68.4 $\pm {\scriptstyle{2.8}}$                           & 47.8 $\pm {\scriptstyle{3.1}}$                           & 48.6 $\pm {\scriptstyle{6.3}}$                           \\
GestureLM \small with only positional& 48.9 $\pm {\scriptstyle{16.1}}$         & 29.3 $\pm {\scriptstyle{24.3}}$         & 50.2 $\pm {\scriptstyle{33.2}}$                          & 41.9 $\pm {\scriptstyle{37.1}}$                          & 34.9 $\pm {\scriptstyle{15.0}}$                          & 26.7 $\pm {\scriptstyle{22.8}}$                          \\ \hline
GestureLM (Ours)                     & \textbf{61.2 $\pm {\scriptstyle{1.5}}$} & \textbf{51.1 $\pm {\scriptstyle{1.7}}$} & \textbf{74.8}$\pm {\scriptstyle{2.6}}$ & \textbf{70.4}$\pm {\scriptstyle{3.1}}$ & \textbf{52.8} $\pm {\scriptstyle{1.7}}$ & \textbf{52.2} $\pm {\scriptstyle{4.3}}$ \\ \hline
\end{tabular}}
\caption{Validating the impact of gesture embeddings through adversarial evaluation. }
\label{tab:sanity}
\end{table*}

\subsection{Comparison with Existing Approaches}

This section addresses two key questions. \textit{First}, how do models that incorporate gesture information compare with Text-only baseline model across the three tasks. \textit{Second}, we evaluate how our approach to integrate gestures with language models compares to prior works that also model language and gestures jointly. We refer to our modeling pipeline as GestureLM in the tables.

We compare our approach with \citet{xu2023spontaneous}, who introduce a grid-based tokenization scheme and a Mixed Modal architecture to integrate gestures with language models. To adapt this approach to our dataset, we apply a 3D-to-2D projection to obtain grid-based tokens and evaluate their modeling pipeline on our tasks. To enable a more direct comparison and better understand the impact of different gesture tokenization strategies, we also compare with their tokenization scheme integrated into our GestureLM pipeline.  Additionally, we compare our approach to GesTran \cite{xu2024llm} in how they encode gestures. They augment the language model's vocabulary with VQ-VAE codebook indices for co-speech gesture generation, however, they do not leverage the learned motion information in the codebook embeddings. In contrast, our approach utilizes gesture embeddings, which inherently capture motion information, and maps them into the language space. More details regarding implementation are provided in the Appendix \ref{appendix:implementation}. 

Overall, the results indicate that incorporating gestures alongside text enhances the model's performance in spoken discourse modeling. This improvement is particularly pronounced in terms of the F1 score, with an average gain of 4.8\% across all three tasks. We also find that gestures play an important role in the prediction of markers that are less frequently used and located in the long tail of the distribution as seen in Figure \ref{fig:cnf}. This is especially relevant and important to capture given that the rare words often convey more specific meaning than frequent ones like \emph{and}. A detailed per-class performance analysis is presented in the Error Analysis section~\ref{sec:error_analysis} to provide further insights. We also find that using learned gesture embeddings through VQ-VAE-based tokenization yields better performance compared to existing gesture encoding approaches.

\subsection{Validating the Impact of Gestures via Adversarial Evaluation}

Inspired by multimodal translation studies analyzing the necessity of visual input for multimodal translation \cite{wu2021good}, we conduct a similar analysis by replacing pre-trained gesture embeddings with adversarial inputs such as random vectors from a normal distribution simulating uninformative gesture representations. Additionally, we experiment with using only positional embeddings while setting gesture embeddings to zero. This helps determine whether the performance gains stem from gesture duration, which may correlate with word length rather than semantic content.

From Table \ref{tab:sanity}, we observe that for all three tasks, using pre-trained gesture embeddings consistently improves F1 scores compared to random vectors or positional embeddings alone. These findings suggest that pre-trained gesture embeddings provide meaningful information rather than acting as a form of regularization. 
We believe that such adversarial evaluation is a crucial step for future research to ensure that the added modality genuinely contributes to language modeling. 






\subsection{Ablations}

\subsubsection{Masking \% during feature alignment}

We select the masking percentage in the feature alignment stage based on the lowest validation loss from the masked language modeling objective. A lower loss reflects better-aligned gesture embeddings, leading to improved language modeling. Table \ref{tab:ppl} shows the relationship between masking percentage and validation loss. As expected, the loss increases with higher masking percentages, but also below 30\%.

\subsubsection{Model Components}

We also ablate the design choices in our modeling pipeline to understand their effect on the task performance. In Table \ref{tab:ablation}, we first compare the use of absolute positional embeddings with relative ones and observe an improvement in model performance across all three tasks. We also test the model without the feature alignment stage, that is, without mapping the representation spaces, and find a significant drop in performance, highlighting the importance of this stage in effectively fusing the two modalities.

\begin{table*}[t]
\centering
\resizebox{\textwidth}{!}{%
\begin{tabular}{lcccccc}
\hline
 & \multicolumn{2}{c}{\textbf{Discourse}} & \multicolumn{2}{c}{\textbf{Quantifer}} & \multicolumn{2}{c}{\textbf{Stance}} \\ \hline
 & Acc (\%)              & F1             & Acc (\%)              & F1             & Acc (\%)            & F1            \\ \hline
GestureLM &
  \textbf{61.2 $\pm {\scriptstyle{1.5}}$} &
  \textbf{51.1 $\pm {\scriptstyle{1.7}}$} &
  \textbf{74.8 $\pm {\scriptstyle{2.6}}$} &
  \textbf{70.4 $\pm {\scriptstyle{3.1}}$} &
  \textbf{52.8} $\pm {\scriptstyle{1.7}}$ & \textbf{52.2} $\pm {\scriptstyle{4.3}}$ \\ 
GestureLM \small w/o relative positional encoding &
  59.6 $\pm {\scriptstyle{2.3}}$ &
  47.6 $\pm {\scriptstyle{2.9}}$ &
  73.8 $\pm {\scriptstyle{6.7}}$ &
  69.1 $\pm {\scriptstyle{3.2}}$ &
  50.6 $\pm {\scriptstyle{3.9}}$ &
  50.8 $\pm {\scriptstyle{3.9}}$ \\
GestureLM \small w/o feature alignment &
  60.2 $\pm {\scriptstyle{4.1}}$ &
  46.7 $\pm {\scriptstyle{5.4}}$ &
  68.8 $\pm {\scriptstyle{4.6}}$ &
  65.1 $\pm {\scriptstyle{5.5}}$ &
  43.0 $\pm {\scriptstyle{34.7}}$ &
  34.7 $\pm {\scriptstyle{13.8}}$ \\ \hline
\end{tabular}
}
\caption{Ablation w.r.t Model Components}
\label{tab:ablation}
\end{table*}

\begin{table}[t]
\centering
\small
\begin{tabular}{lcccc}
\hline
Masking percentage & 10\% & 30\% & 50\% & 80\%  \\ \hline
loss ($\downarrow$)              & 1.4 & \textbf{0.3}  & 0.4 & 4.9 \\ \hline
\end{tabular}
\caption{Masking percentage vs validation loss from the feature alignment stage}
\label{tab:ppl}
\end{table}


\subsection{Error Analysis}
\label{sec:error_analysis}
\begin{figure}[h!]
\centering
    \begin{subfigure}{0.48\textwidth}
    \centering
\includegraphics[width=0.9\textwidth]{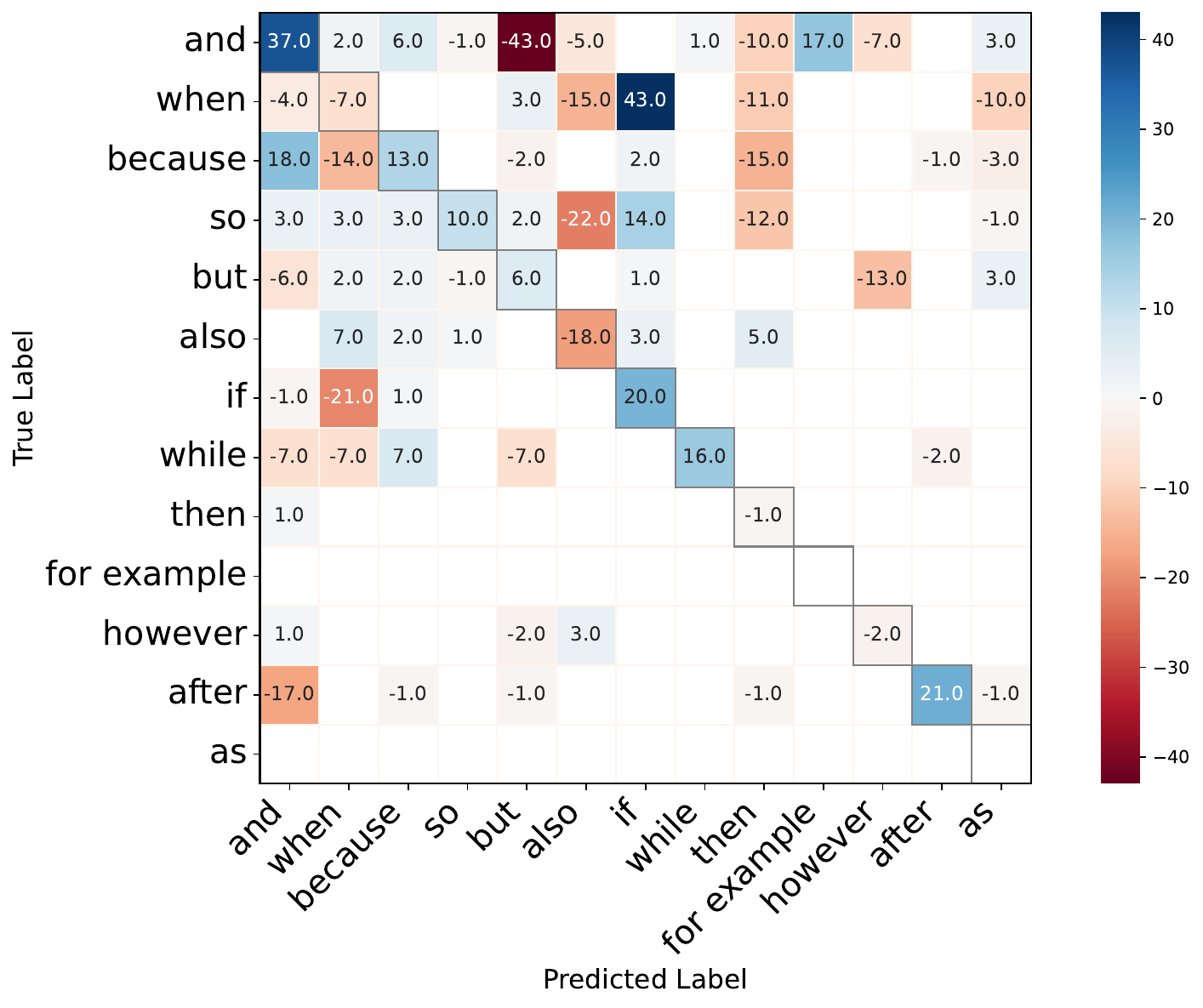}
    \caption{Discourse Connective}
    \label{fig:discourse_cnf}
    \end{subfigure}
    \vfill
    \vspace{0.7pt}
    \begin{subfigure}{0.23\textwidth}
    \centering
    \includegraphics[width=\textwidth]{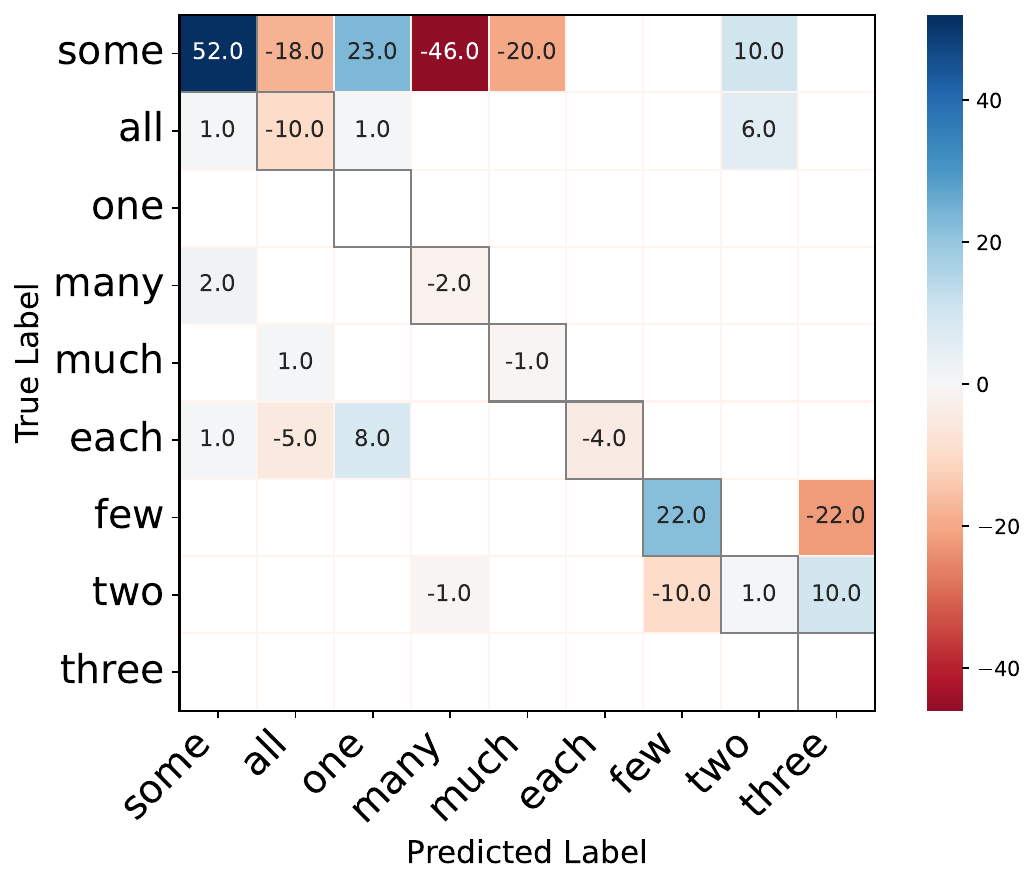}
    \caption{Quantifier}
    \label{fig:quantifier_cnf}
    \end{subfigure}
    ~
    \begin{subfigure}{0.23\textwidth} 
\includegraphics[width=\textwidth]{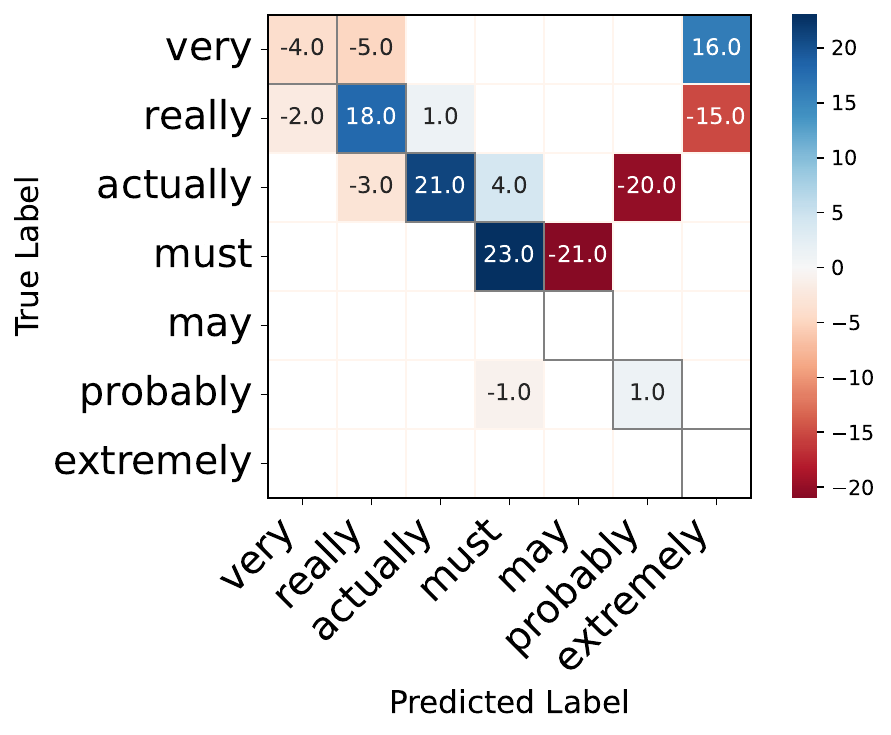}
    \caption{Stance Marker}
    \label{fig:stance_cnf}
    \end{subfigure}
    \caption{Relative Confusion Matrices comparing GestureLM and Text-only baseline. The matrix highlights differences in class-wise predictions, with red indicating more Text-only predictions and blue signifying more by GestureLM.}
    \label{fig:cnf}
\end{figure}

\begin{figure}[h!]
\centering
    \begin{subfigure}{0.45\textwidth}
    \centering
    \includegraphics[width=\linewidth]{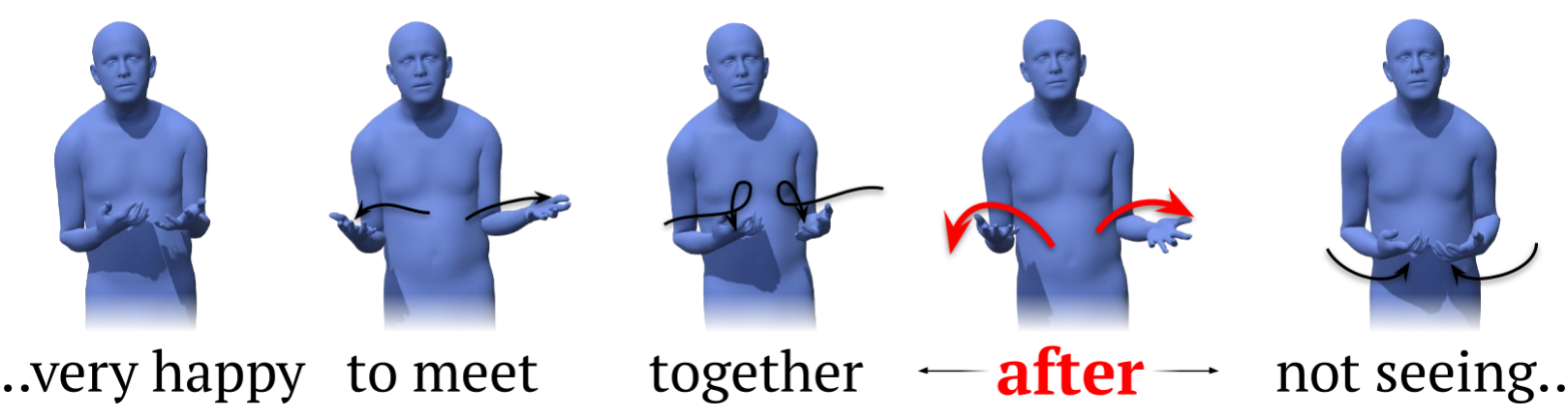}
    \caption{Gesture occurring in the lateral axis near the temporal discourse connective \textit{after}.}
    \label{subfig:discourse-carlos-95}		
    \end{subfigure}
    
    \vfill
    \vspace{0.5pt}
    
    \begin{subfigure}{0.45\textwidth}
    \centering
    \includegraphics[width=\linewidth]{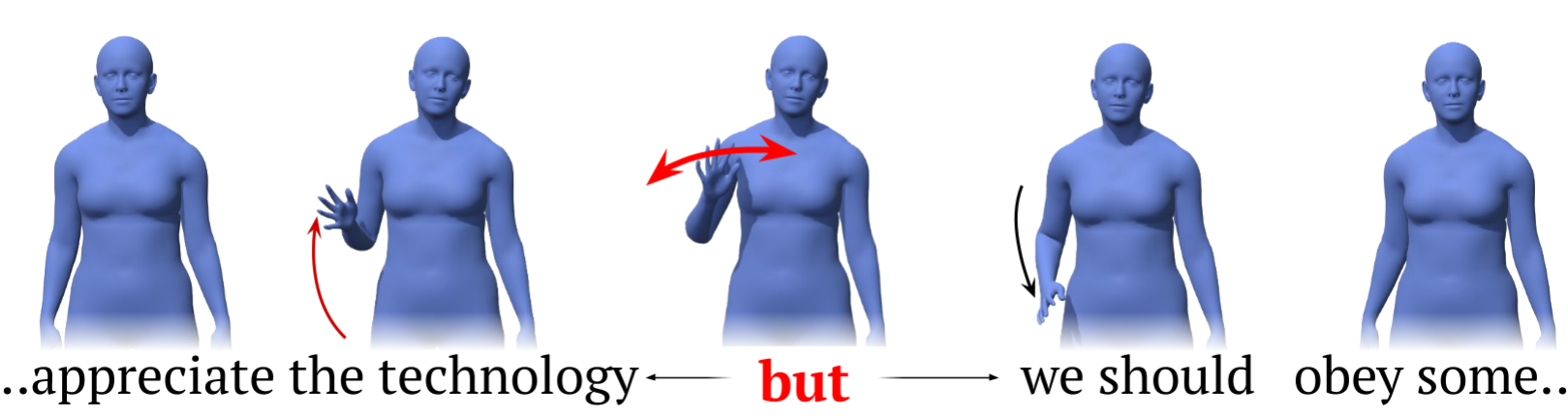}
    \caption{Raised hand gesture to indicate contrast using discourse connective \textit{but}.}
    \label{subfig:discourse-miranda-111}
    \end{subfigure}
    \vfill
    \vspace{0.5pt}

    \begin{subfigure}{0.45\textwidth}
    \centering
    \includegraphics[width=\linewidth]{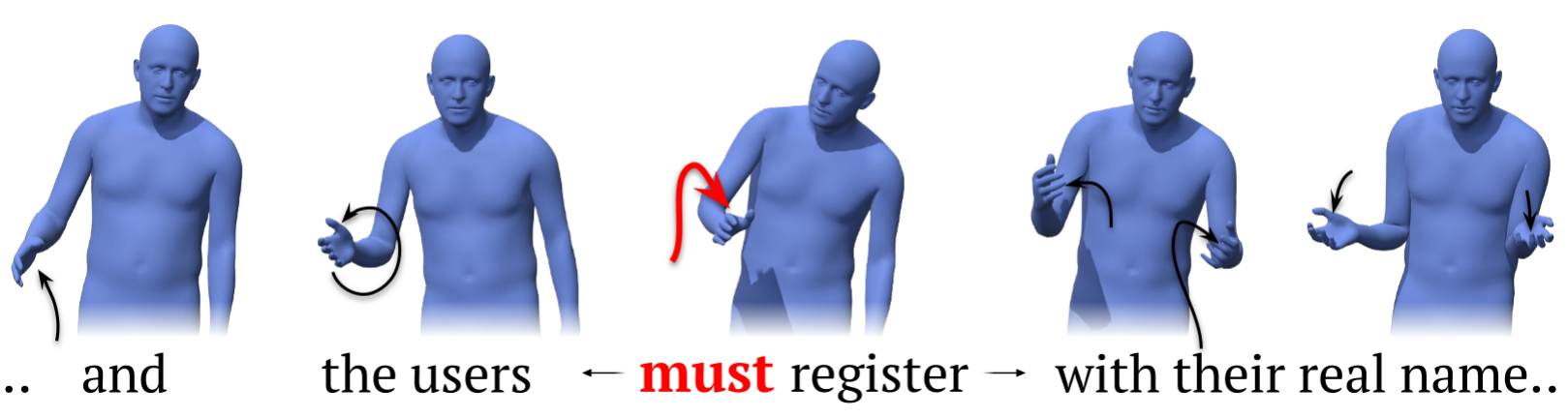}
    \caption{Palm-down gesture occurring during the utterance of \textit{must}, an epistemic stance marker conveying assertiveness.}
    \label{subfig:stance-scott-87}		
    \end{subfigure}
    \vfill
    \vspace{0.5pt}
    \label{fig:examples}
    \caption{Some samples where GestureLM performs better than Text-only model. We see that the semantic gestures co-occur with the spoken discourse markers, potentially leading to improved prediction performance. See the text for detailed description and more examples are shown in the Appendix \ref{appendix:examples}.}
	
\end{figure}

Overall, we find that gesture incorporation enhances the performance of language models, especially for underrepresented markers in the training data. However, our error analysis also highlights certain challenges, especially in disambiguating confusable markers. To further investigate this, we analyze the class-wise task performance of the models using a relative confusion matrix \cite{pomme2022relative}, shown in Figure \ref{fig:cnf}. This matrix captures the difference in predictions between the two comparison models which in our case is GestureLM vs.~Text-only baseline. In terms of interpretation, a stronger red intensity indicates more predictions by the Text-only model for a given cell, while a stronger blue intensity signifies more predictions by the GestureLM. %
Note that the rows are arranged in descending order of marker frequency in test samples, from top to bottom. Rows and columns with minimal difference are omitted for space considerations, and the complete confusion matrices are available in the Appendix \ref{appendix:cnf}.


In the discourse connective prediction task, we observe improved performance particularly for temporal discourse relations such as \textit{after} and \textit{while}. This improvement can be attributed to the presence of gestures associated with these connectives near their utterance as seen in the samples in the Figure \ref{subfig:discourse-carlos-95}. These gestures likely help to disambiguate these connectives from more commonly used ones, such as \textit{and} (refer to the training distribution in Figure \ref{fig:data}), which the Text-only model often confuses temporal connectives with, as seen in Figure~\ref{fig:discourse_cnf} (red cells in the \textit{and} column). 
Linguistic research further supports that speakers use spatial gestures to express temporal concepts such as \textit{after} and \textit{before} \cite{casasanto2012hands}.
%
Similarly, for contrastive connectives like \textit{but}, gestures such as a raised hand during the utterance of \textit{but} (shown in the Figure \ref{subfig:discourse-miranda-111}) potentially helps GestureLM distinguish it from \textit{and}.


%
Similarly, in the quantifier task, we see in Figure~\ref{fig:quantifier_cnf} that incorporating gestures helps with underrepresented classes such as \textit{few} and \textit{some}. Particularly, in the case of \textit{some}, while the Text-only models confuse it with higher represented classes such as \textit{all} and \textit{much}, the gesture models tend to confuse it with numbers such as \textit{two} or \textit{one}. One possible explanation is that similar wrist movements can used to represent both types of quantifiers, as the specific finger joint movements that convey numerical information are not captured. Future work could incorporate data from finger joints to better distinguish between these classes. 



In stance marker prediction, incorporating gestures helps disambiguate \textit{must} with \textit{may}, which Text-only models often confuse. \textit{Must}, which conveys  assertiveness belongs to epistemic stance which is found to co-occur with gestural cues \cite{andries2023multimodal,marrese2021grammar}.  For instance, in Figure~\ref{subfig:stance-scott-87}, we see the speaker performing a palm-down gesture which could help the gesture-enhanced models in better disambiguation. Additionally, we see in Figure~\ref{fig:stance_cnf} that both models tend to confuse the most commonly occurring class, \textit{really}, with \textit{very} and \textit{extremely}. Unlike previous tasks, these mistakes are acceptable as the words are often interchangeable. More examples are in Appendix \ref{appendix:examples}



\section{Conclusion}
In this work, we presented a framework that integrates gestural cues into language models to enhance the modeling of spoken discourse. To evaluate our approach, we constructed linguistically grounded text infilling tasks focused on three types of spoken discourse markers: discourse connectives, quantifiers and stance markers. Our results demonstrate that incorporating gestural information improves the accuracy of spoken discourse marker prediction in language models. We view this work as a starting step toward multimodal spoken discourse modeling in language models. Future research should investigate whether gestures can also contribute in cases where meaning is not explicitly conveyed through language but is instead expressed solely through gesture (e.g., implicit coherence relations), which would require annotated data for evaluation.


\section{Limitations}
First, our model captures only the upper-body joints up to the wrist, excluding finger joint movements. As a result, gestures relying on finger joints, such as distinguishing between \textit{two} and \textit{three} cannot be disambiguated. Future work will incorporate these finer-grained features to improve the model's ability to capture subtle gesture variations. Second, our approach is currently designed for encoder-only MLM-based models, as they are generally used for token classification. In future work, we aim to explore how gesture-enhanced models can be helpful in decoder-based models that are trained for next-token prediction. Third, the dataset used in this study captures a specific group of speakers and communicative contexts, which may not fully reflect the diversity of gesture used across different cultures, languages, and interaction settings. Lastly, to represent body joints in motion, we use 6D rotation representation in this work. However, the choice of other representations, such as joint positions, acceleration etc., could vary the performance of spoken language modeling. Additionally, many real-world scenarios lack 3D motion capture data and instead rely on 2D video, which may reduce gesture encoding accuracy and limit the effectiveness of integrating gestures into language models.

\section{Ethical Statement}

This work uses publicly available dataset and uses only the raw 3D joint data for gesture modeling, without utilizing personally identifiable information or video recordings of individuals. However, we acknowledge potential ethical concerns, as gesture data may unintentionally reveal personal attributes, such as gender or physical characteristics, which could pose privacy risks if misused for unintended inferences.


\section*{Acknowledgments}
This work was supported by the Deutsche Forschungsgemeinschaft, Funder Id: \url{http://dx.doi.org/10.13039/501100001659}, Grant Number: SFB1102: Information Density and Linguistic Encoding. M. Hamza Mughal was funded by the Deutsche Forschungsgemeinschaft (DFG, German Research Foundation) - GRK 2853/1 “Neuroexplicit Models of Language, Vision, and Action” - project number 471607914.

\bibliography{custom}

\appendix

\section{Appendix}

\subsection{Implementation Details (Cont.)}
\label{appendix:implementation}
The three main steps in our approach are gesture tokenization, followed by feature alignment, and then low-rank adaptation. 
In the gesture tokenizer, the transformer encoder and decoder consist of 8 layers, with each layer containing 4 attention heads.
The whole pipeline is trained for 57 epochs with AdamW optimizer with a learning rate of $3e-5$.
For the feature alignment step, we have a total of 1.8M trainable parameters (total = 126.6M). We use the AdamW optimizer with a learning rate of $1e-3$, a cosine scheduler with a warmup ratio of 0.03, and train for 20 epochs. Early stopping is based on validation loss, with a batch size of 32. For the fine-tuning stage, we have a total of 4.7M trainable parameters (LoRA), for both Text-only model and GestureLM. For all tasks and settings, we train with a learning rate of $1e-3$ with a cosine scheduler with a warmup ratio of 0.03, and 10 epochs. The best model is chosen based on validation F1, with a batch size of 16. We use the AdamW optimizer with a weight decay of 1e-3 for this step. The hyperparameters were obtained via manual tuning. 

For the comparison approaches, we adapt the Mixed Modal method from \cite{xu2023spontaneous} for the BEAT2 dataset. Specifically, we convert the tokens from 3D to 2D, and to ensure a fairer comparison, we use pre-trained RoBERTa text embeddings instead of training from scratch. We adopt the concatenation fusion approach from their paper, as it was shown to yield the lowest validation loss. Since the main contribution of their paper is the tokenization scheme, we compare their tokenization method with ours (GestureLM with grid-based tokens, as shown in Table \ref{tab:main}). For this comparison, we create an Embedding Layer with grid token indices and train these embeddings during our feature alignment stage, as they are randomly initialized. A similar approach is used when incorporating the codebook indices similar to \cite{xu2024llm}. For training, we use the same parameters mentioned in the paragraph above.

We run all our experiments on a single NVIDIA Quadro RTX 8000 (48 GB GDDR6) GPU. For implementation we use PyTorch \footnote{\url{https://pytorch.org/}}, Roberta weights from HuggingFace transformers \footnote{\url{https://huggingface.co/}} and AdapterHub for LoRa implementation \footnote{\url{https://docs.adapterhub.ml/index.html}}

\subsection{Losses from the gesture tokenization}
\label{appendix:gesture_tokenization}
Our tokenization pipeline is trained using VQ-VAE for motion reconstruction.
Therefore, we apply standard MSE loss on the decoder output $\mathbf{\hat{x}}$ in 6D rotational representation and also apply the reconstruction losses by converting $\mathbf{\hat{x}}$ to axis-angle representation and joint positions.
We also apply Geodesic Loss on rotation matrices.
Along with that, we apply MSE losses on velocity and accelaration of the motion.
The codebook loss includes the VQ objective and the commitment term ~\cite{van2017neural}.
%
\begin{align}
    \mathcal{L}_{rec} =& \mathcal{L}_{6D} + 
    \mathcal{L}_{axis-angle} + \nonumber
    \\ &
    \mathcal{L}_{joint-pos} + \mathcal{L}_{rot}\\
    \mathcal{L}_{codebook} =& {|| \textrm{sg} [\mathbf{z}] - \mathbf{z}^q ||}^2 + \beta {|| \mathbf{z} - \textrm{sg}[\mathbf{z}^q] ||}^2\\
    \mathcal{L}_{\textrm{VQ-VAE}} =& \mathcal{L}_{rec}(\mathbf{x}, \mathbf{\hat{x}}) + \mathcal{L}_{codebook} +  
    \nonumber \\&
    \mathcal{L}_{vel}(\mathbf{x'}, \mathbf{\hat{x}'}) + \mathcal{L}_{acc}(\mathbf{x''}, \mathbf{\hat{x}''}) 
\end{align}
Here, $\textrm{sg}$ is the stop gradient operation and the weight of commitment term $\beta=0.25$.
%





\subsection{Obtaining Gesture Positional Embedding}
\label{appendix:feature}

To obtain gesture positions, we first determine the timestamp of each text token using Whisper. Next, we extract the gesture tokens that fall within this time window. For example, if the word \textit{really} is spoken over 0.5 seconds and each gesture token corresponds to 4 frames at a frame rate of 15 fps, we get approximately 2 (1.875) gesture tokens for this word. We then assign the text token’s position to all the extracted gesture tokens associated with it. Since Whisper only provides time durations for whole words but not individual tokens, we distribute the total duration of multi-token words proportionally based on the number of characters in each token.

\begin{figure*}[h!]
\centering
    \begin{subfigure}{0.6\textwidth}
    \centering
    \includegraphics[width=\linewidth]{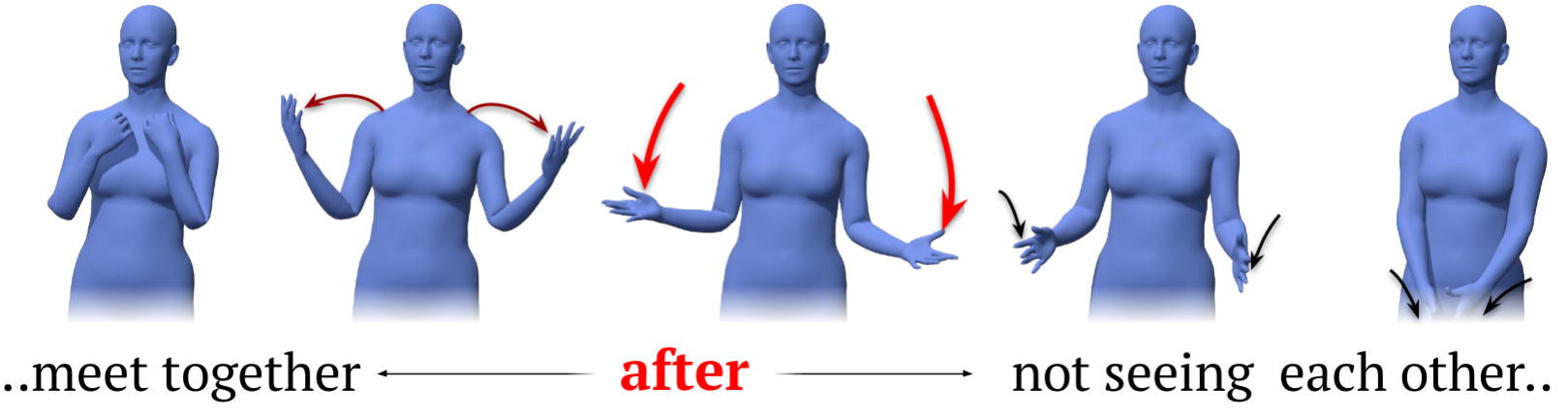}
    \caption{Another example of spatial gesture occurring near the temporal discourse connective \textit{after}. The Text-only model predicts \textit{but}, while GestureLM \textbf{correctly} predicts \textit{after}.}
    \label{subfig:discourse-yinqang-95}
    \end{subfigure}
    \vfill
    \vspace{0.5pt}
    
    \begin{subfigure}{0.6\textwidth}
    \centering
    \includegraphics[width=\linewidth]{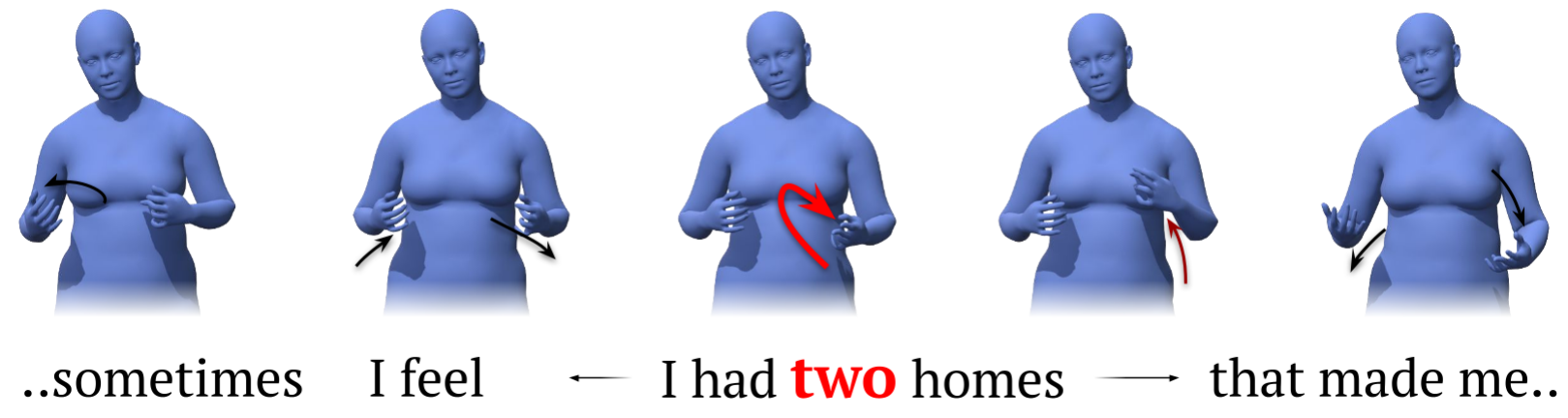}
    \caption{In this example, the speaker lifts the left hand to depict \textit{two} using fingers. The Text-only model predicts \textit{many}, while GestureLM \textbf{correctly} predicts \textit{two}.} 
    
    \label{subfig:quantifier-sophie-6-6}
    \end{subfigure}
    \vfill
    \begin{subfigure}{0.6\textwidth}
        \centering
		\includegraphics[width=\linewidth]{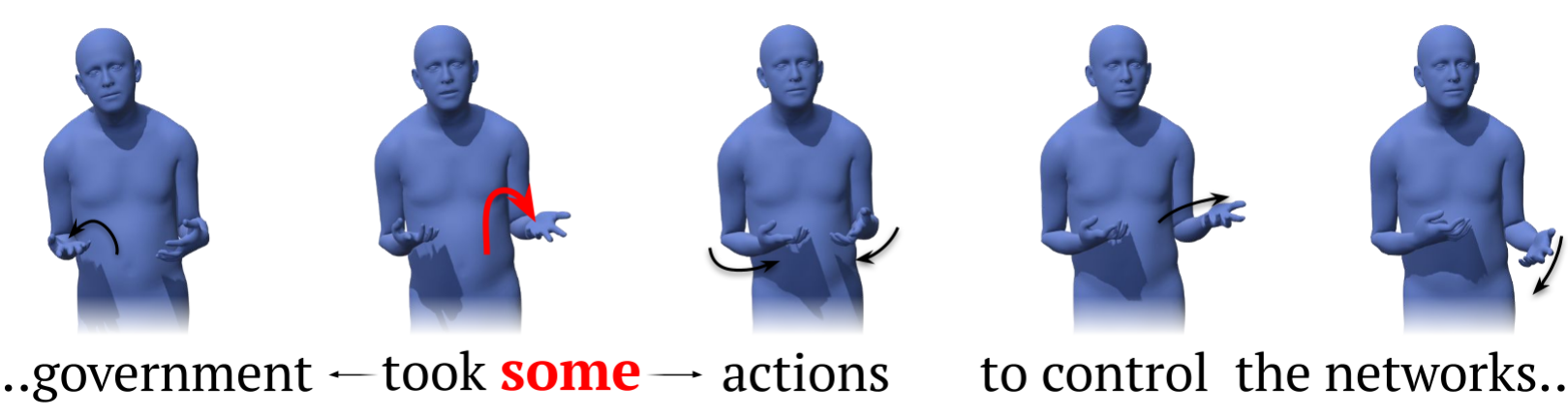}
		\caption{Similar to \ref{subfig:quantifier-sophie-6-6}, the speaker flicks the left hand to depict \textit{some}. The Text-only model predicts \textit{many}, while GestureLM \textbf{incorrectly} predicts \textit{two}. Incorporating finger joints can help GestureLM for better disambiguation in both examples.}
		\label{subfig:quantifier-carlos-87}		
	\end{subfigure}

    \vfill
    \begin{subfigure}{0.6\textwidth}
        \centering
		\includegraphics[width=\linewidth]{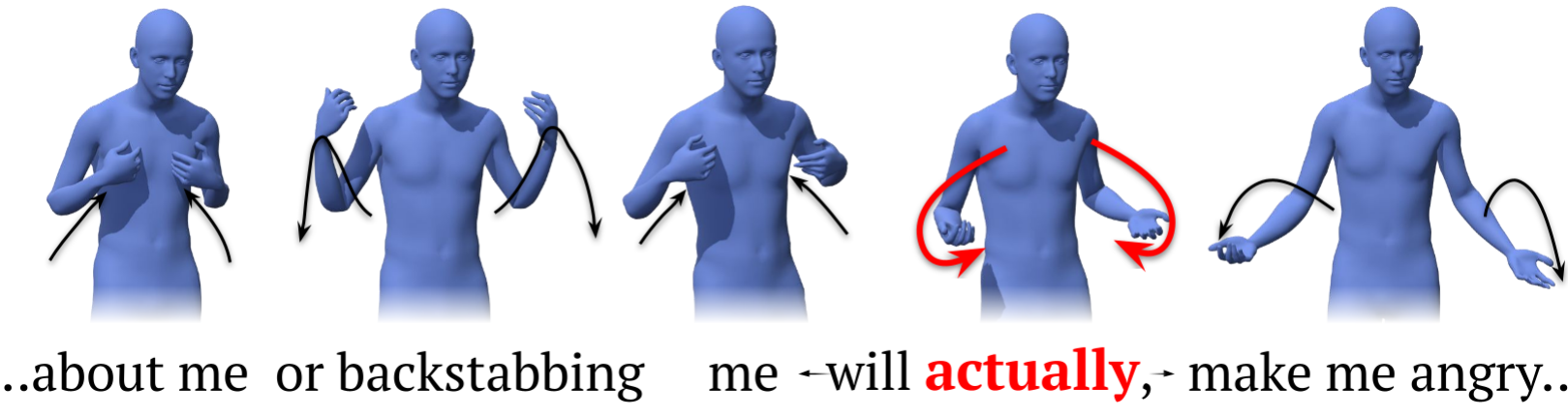}
		\caption{Gesture occurring near the epistemic stance marker \textit{actually}. The Text-only model predicts \textit{probably}, while GestureLM \textbf{incorrectly} predicts \textit{really}. However, in the case of GestureLM this mistake is acceptable as opposed to the Text-model, this further signifies the importance of incorporating non-verbal cues.}
		\label{subfig:stance-lu-73}		
	\end{subfigure}

    \vfill
    \begin{subfigure}{0.6\textwidth}
        \centering
		\includegraphics[width=\linewidth]{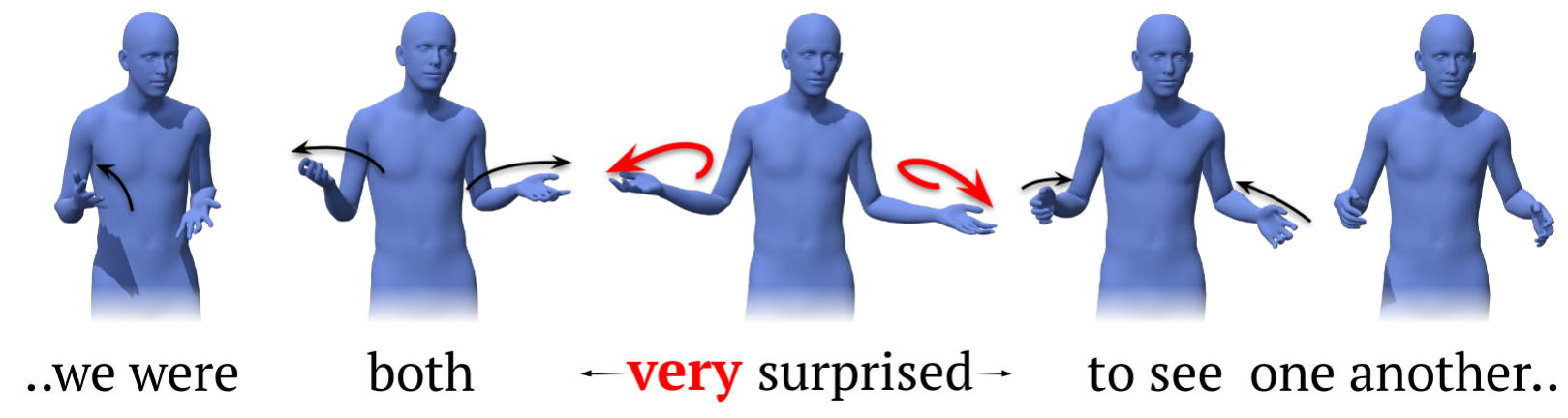}
		\caption{Another example of gesture occurring near the epistemic stance marker \textit{very}. Both model \textbf{incorrectly} predicts \textit{really} in this case. This gesture is similar in motion to \ref{subfig:stance-lu-73}	}
		\label{subfig:stance-lu-95}		
	\end{subfigure}

    \vfill
    \begin{subfigure}{0.6\textwidth}
        \centering
		\includegraphics[width=\linewidth]{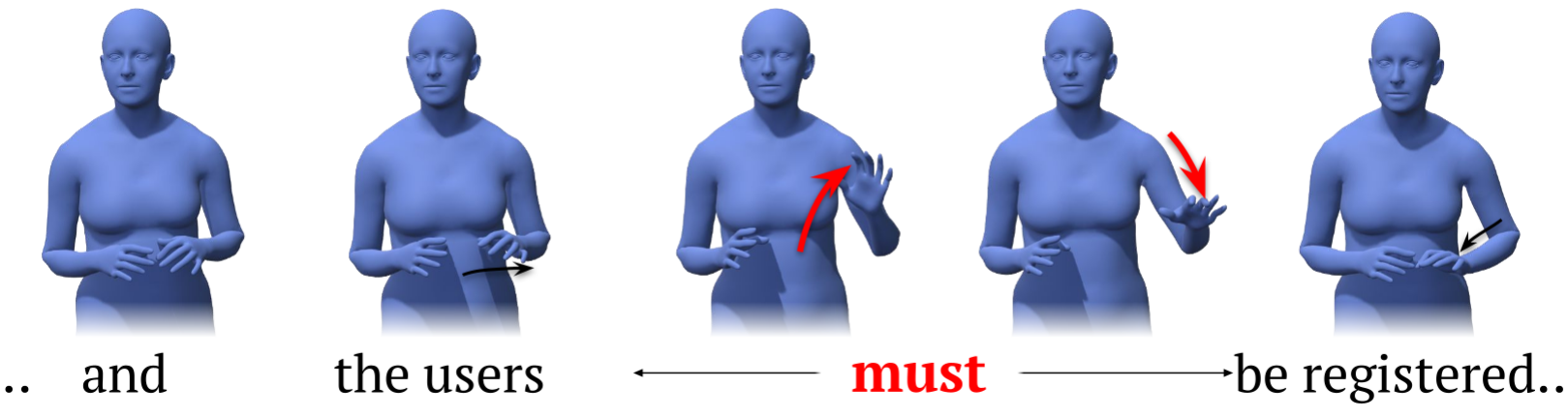}
		\caption{Another example of palm down gesture happening at stance marker \textit{must} similar to one displayed in the main text.}
		\label{subfig:stance-ayana-87}		
	\end{subfigure}
    
	\caption{Examples
    }
	\label{fig:appendix_examples}
    \vspace{-10pt}
\end{figure*}

\subsection{More examples}
\label{appendix:examples}
Figure \ref{fig:appendix_examples} presents additional examples of gesture sequences accompanied by a detailed explanation in the caption.

\subsection{Confusion Matrix}
\label{appendix:cnf}
\begin{figure*}[h!]
	\centering
    \begin{subfigure}{0.9\textwidth}
        \centering
\includegraphics[width=\linewidth]{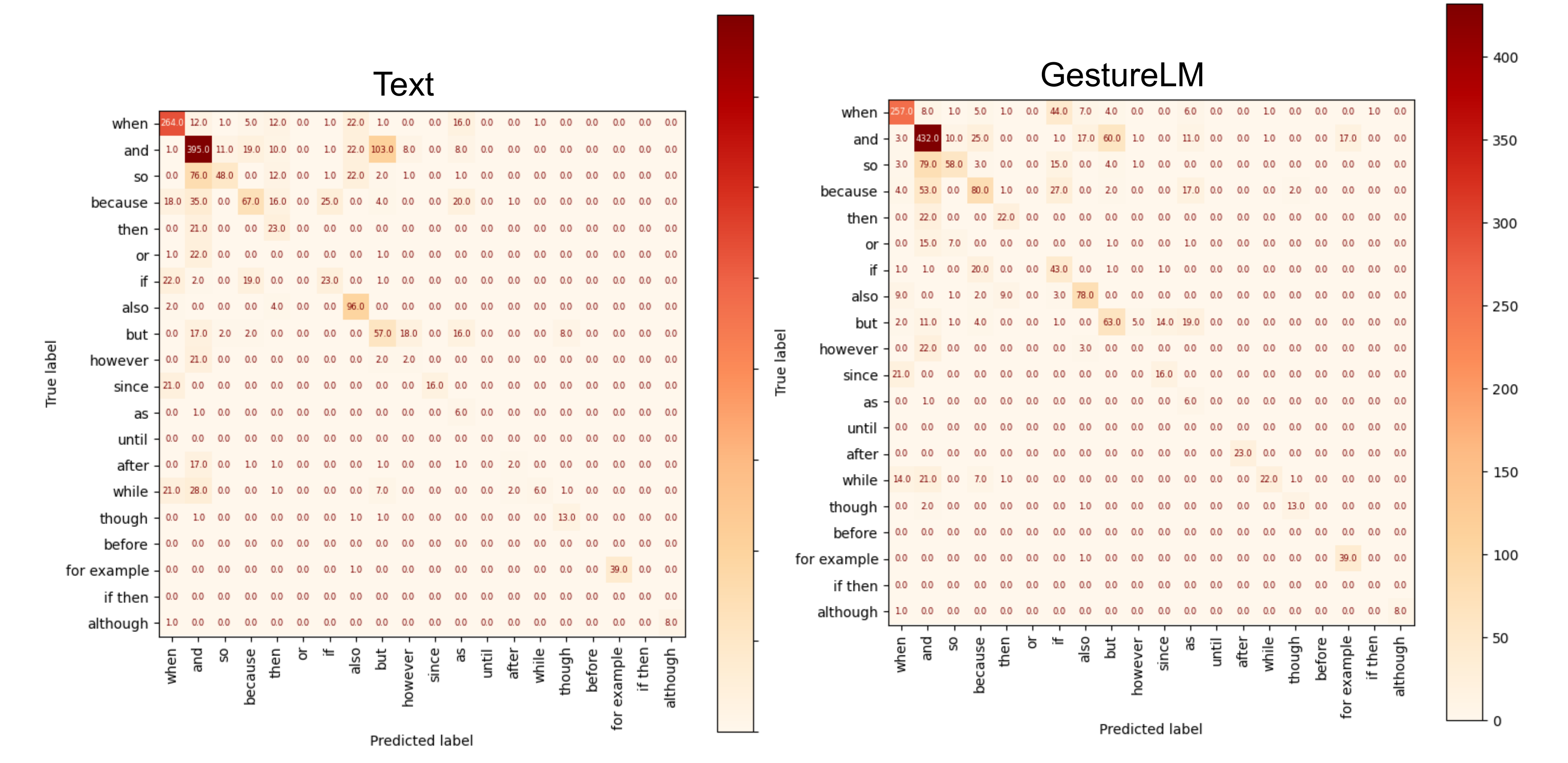}
		\caption{}
    \label{fig:discourse_full_cnf}		
	\end{subfigure}
    \vfill
    \vspace{0.5pt}
    \begin{subfigure}{0.9\textwidth}
    \centering
\includegraphics[width=\linewidth]{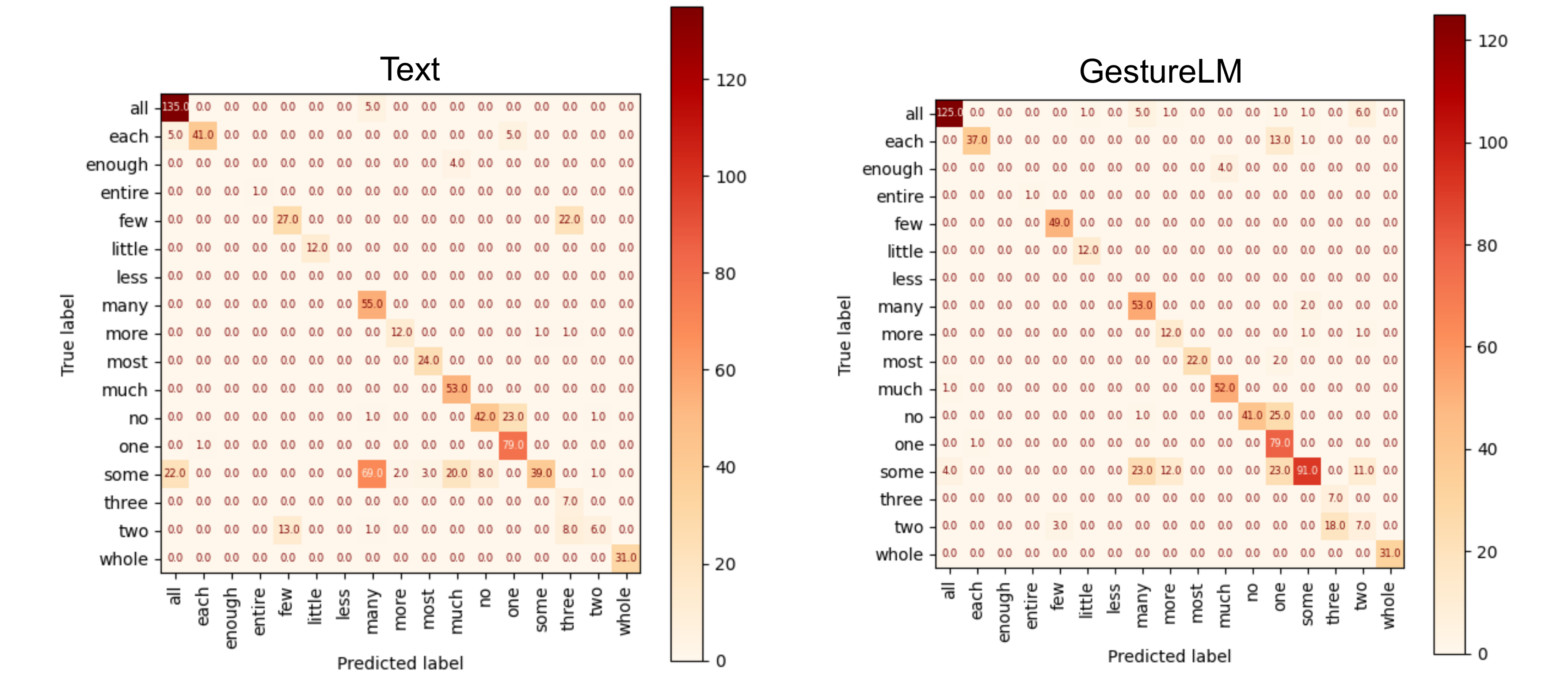}
		\caption{} \label{fig:quantifier_full_cnf}
	\end{subfigure}
    \vfill
    \vspace{0.5pt}
    \begin{subfigure}{0.9\textwidth}
        \centering
		\includegraphics[width=\linewidth]{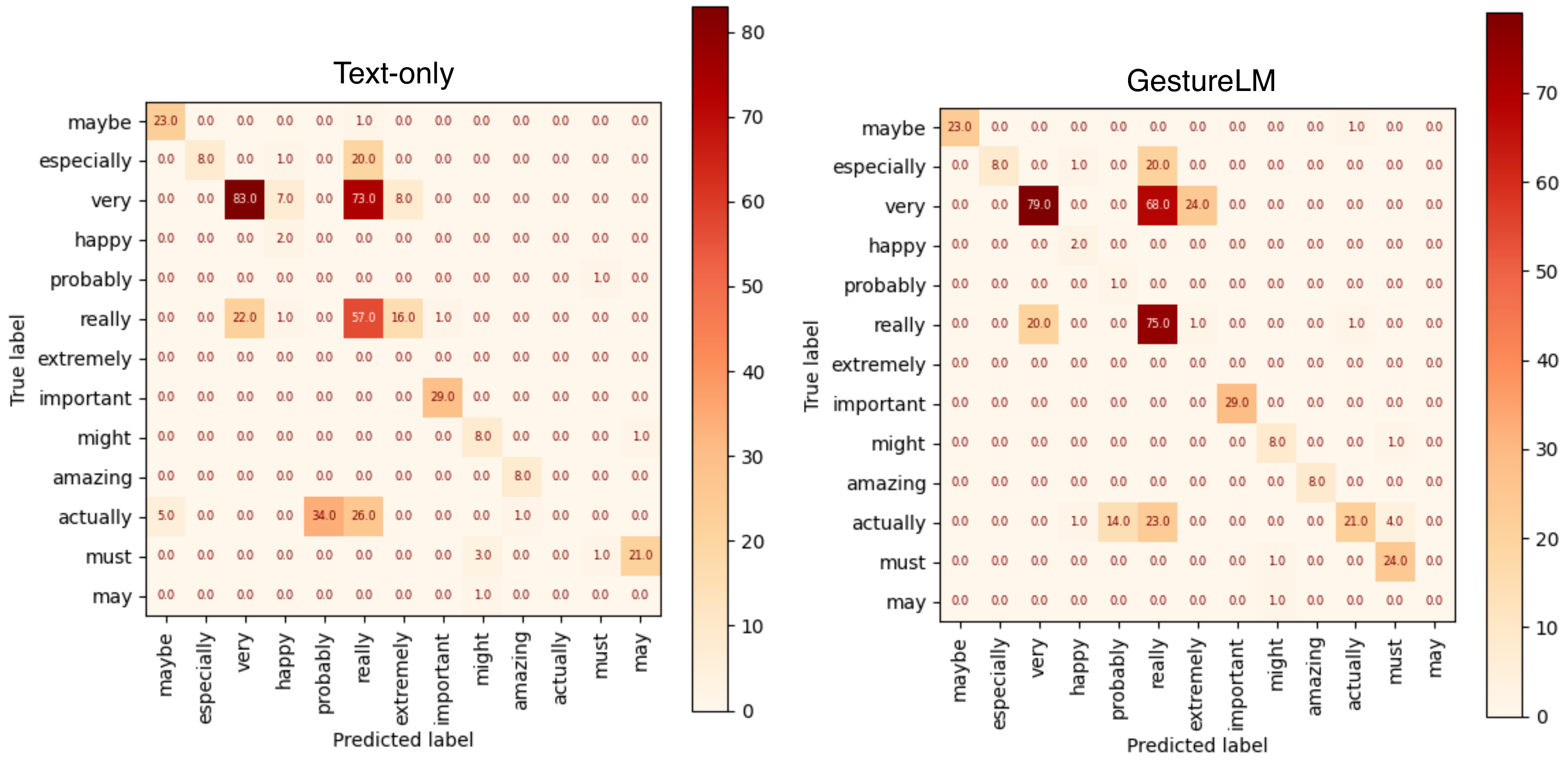}
		\caption{}
		\label{fig:stance_full_cnf}		
	\end{subfigure}
	\caption{Confusion matrices
    }
	\label{fig:full_cnf}
    \vspace{-10pt}
\end{figure*}

Figure \ref{fig:discourse_full_cnf}, \ref{fig:quantifier_full_cnf} and \ref{fig:stance_full_cnf} illustrates the full confusion matrices for discourse connectives, quantifiers and stance markers respecitively.

\end{document}